\documentclass[runningheads]{llncs}
\usepackage{graphicx}
\graphicspath{ {fig/} }
\usepackage{tabularx}
\usepackage{algorithmic}
\usepackage{algorithm}
\usepackage{rotating}
\usepackage{amsmath}
\begin{document}
\title{S-Extension Patch: A simple and efficient way to extend an object detection model\thanks{Supported by AV DEVS Solutions Pvt. Ltd.}}
%
%\titlerunning{Abbreviated paper title}
% If the paper title is too long for the running head, you can set
% an abbreviated paper title here
%
\author{Dishant Parikh\inst{1}}
\authorrunning{D. Parikh}
% First names are abbreviated in the running head.
% If there are more than two authors, 'et al.' is used.
%
\institute{AV DEVS Solutions Pvt. Ltd., Gujarat, India \\ 
\email{dishant30899@gmail.com}\\
\url{https://www.avdevs.com}}
\maketitle              % typeset the header of the contribution
\begin{abstract}
While building convolutional network-based systems, the toll it takes to train the network is something that cannot be ignored. In cases where we need to append additional capabilities to the existing model, the attention immediately goes towards retraining techniques. In this paper, I show how to leverage knowledge about the dataset to append the class faster while maintaining the speed of inference as well as the accuracies; while reducing the amount of time and data required. The method can extend a class in the existing object detection model in 1/10th of the time compared to the other existing methods. S-Extension patch not only offers faster training but also speed and ease of adaptation, as it can be appended to any existing system, given it fulfills the similarity threshold condition.

\end{abstract}
\section{Introduction}
With the increase of adaptation to AI infrastructure, the overall need for data and data engineering has increased. When it comes to computer vision, the majority of models take a large amount of data, as well as time for processing the data to train upon. This holds true even for seemingly easy tasks like adding an extension class(es) to an object detection model. To build a network with the same inference power with the extension class enabled, it takes a lot of time in terms of training, data gathering as well as preprocessing (majorly, annotating the dataset). 

When it comes to data, there is some work available to understand and optimize the amount of data required for the particular task. But the work for preprocessing (annotations) and training time will still remain expensive. Even with techniques like Learning without forgetting, the need for preprocessing (annotating) the dataset and the time to train the entire object detection model is still high. When it comes to industrial problems and systems, sometimes the crucial part is the speed of adaptation and updating the system, at least as a placeholder system till the final-efficient model is built. There aren’t many systems in place for doing this, but there are some scenarios in which some information can be leveraged about the dataset to quickly make a system. Here I propose one such system - a similarity-based extension patch. (S-Extension patch) 

The S-Extension Patch is a stand-alone dual-network system that can be used to quickly train and deploy an existing object detection model appended with additional visual capabilities. The dual system consists of a classifier appended to a regular object detection model. The speed of inference is maintained by using a dual-parallel inference technique, which runs both models on different threads sharing the same source data.
The method surpasses other techniques for extending an object detection model in terms of training speed as well as the need for preprocessing data. The S-E patch doesn't require the additional data like Joint training \cite{soltau2014joint}, nor does it require the additional annotations (for the extension classes) or the time to train an entire object detection model, as in Learning without forgetting technique \cite{li2017learning}. The S-E patch, however, has one base condition to be fulfilled, that is, the similarity threshold. The similarity threshold is used to assess the existing model's compatibility with the extension class(es). The S-E patch requires at least one base class for each of the extension classes, which crosses the similarity threshold.

\textbf{Definition of similarity:} 
The average distance between the feature vectors or distance between the feature centroids of two classes is defined as the similarity between the two classes. The distance value is inversely proportional to similarity.

\textbf{Significance of similarity threshold:} 
If any base class A crosses the similarity threshold with the extension class B, then at the time of inference, the model will trigger the region proposal for B, confusing it with A. This confusion is something that can be leveraged in the form of an S-E patch.

For a system that depends on the careful selection of extension classes, the calculation of similarity and decision of the threshold becomes a critical part of the system. For similarity calculation, I went with the content-based image similarity (CBIS) system \cite{gudivada1995content}, \cite{datta2005content}, \cite{eakins1999content}. CBIS helps in understanding the machine's perspective of the dataset, which in turn can be leveraged to our benefit. The knowledge of how the machine is viewing different classes in the dataset could be crucial information in the selection procedure.

\section{Related works}
With the rapid development in deep learning, more powerful tools, which are able to learn semantic, high-level, deeper features, are introduced to address the problems existing in traditional architectures. Deeper and better architectures were introduced, improving the previous algorithms consistently \cite{shin2016deep}. There was one problem though. The computational resources and data preprocessing required in some of the visual tasks like in the work of object detection and segmentation. These problems become an obstacle especially when the task of extending the capabilities of an existing visual system is tackled. A seemingly small task of adding an additional class in the object detection model can be computationally and data-wise extensive. Many techniques were devised to handle this problem and also to look closely into extending a visual system.  

We saw work in fine-tuning the model with respect to the newer requirements, by distilling the knowledge and use of ensemble model \cite{hinton2015distilling} and also by adding units in the neural network for a better performance \cite{wang2017growing}. This gave the introduction to techniques like Learning without forgetting in deep neural networks with the source model and without it by H.Jung \cite{jung2016less} and Zhizhong Li \cite{li2017learning} respectively. Both of which worked quite well than techniques like fine-tuning and joint training \cite{jung2015joint}. Although the need for preprocessing (at least for new visual capabilities) still exists in this technique. Additionally the adaptation of Learning without forgetting as well as other techniques take time and it takes the same amount of time in training a single model too. Although there are cases where this solution is the optimal one, there are cases where this solution can be improved upon.

When we look at the data and the field of content-based image similarity and using the feature vector as a representation of the image, we get a knowledge base that can be leveraged to improve the technique for extending an object detection model. There has been a lot of work in image feature uses in image retrieval systems \cite{ragkhitwetsagul2018picture}, \cite{gudivada1995content}, quality assessments \cite{ding2020image}, and similarity calculations \cite{shnain2017feature}, \cite{plummer2020these}. The technique has been fine-tuned with architectures and methods like PPIS-JOIN \cite{zhang2021ppis} and using histograms as image feature descriptors \cite{wang2020image}. I have combined the same methods with the extension problems of object detectors and conceptualized it as the S-Extension patch. 

\section{Methodology}

\subsection{Similarity threshold check (Compatibility check)}

As explained earlier, the use of similarity threshold is to check the compatibility of the extension class with the base model. For this, I first calculated the similarity of each base class with the extension class and then checked the distances for threshold-based selection. There are three steps to similarity matrix generation. 
\begin{enumerate}
    \item Feature centroid generation.
    \item Calculating similarity by comparing the distances of feature centroids.
    \item Generating similarity matrix by placing the pairwise distance evaluations of the similarity for each class in the dataset.
\end{enumerate}

Feature centroids are generated by fitting an unsupervised clustering algorithm with one cluster and extracting the cluster centers as the feature centroids. The use of clustering algorithms also helps in checking the inertia, i.e., the overall spread of the features in the class, which in turn can be used for checking the overall variety of the class. But this way of checking variety works only if there is no other factor to be taken into account. There are two reasons for computing feature centroids: One, reducing the number of comparisons required. Second, the class should not be checked based on either outliers or extremes that will only hurt the distance computations for similarity.

Once the feature centroids are extracted, the similarity between the classes can be computed by taking the cosine distance (as per equation 1) between the feature centroids of each class. The reason for using cosine distance is that it achieves the highest mMAP score in CBIR systems when used with the feature vector of the average pooling layer. Finally, the matrix is generated by computing the pairwise distances of all classes with all other classes in the dataset.

\begin{equation}
\operatorname{similarity}(A, B)=\frac{A \cdot B}{\|A\| \times\|B\|}=\frac{\sum_{i=1}^{n} A_{i} \times B_{i}}{\sqrt{\sum_{i=1}^{n} A_{i}^{2}} \times \sqrt{\sum_{i=1}^{n} B_{i}^{2}}}
\end{equation}

\subsection{Compatible classes selection}

Once feature centroids for all classes are formed, I calculate the distance between each of the base classes and extension class(es). If the distance is less than the defined similarity threshold then we select those classes and make a classifier for those base classes and the extension class.

\begin{figure}[!t]
\centering
\includegraphics[width = 0.9\columnwidth,scale=0.2]{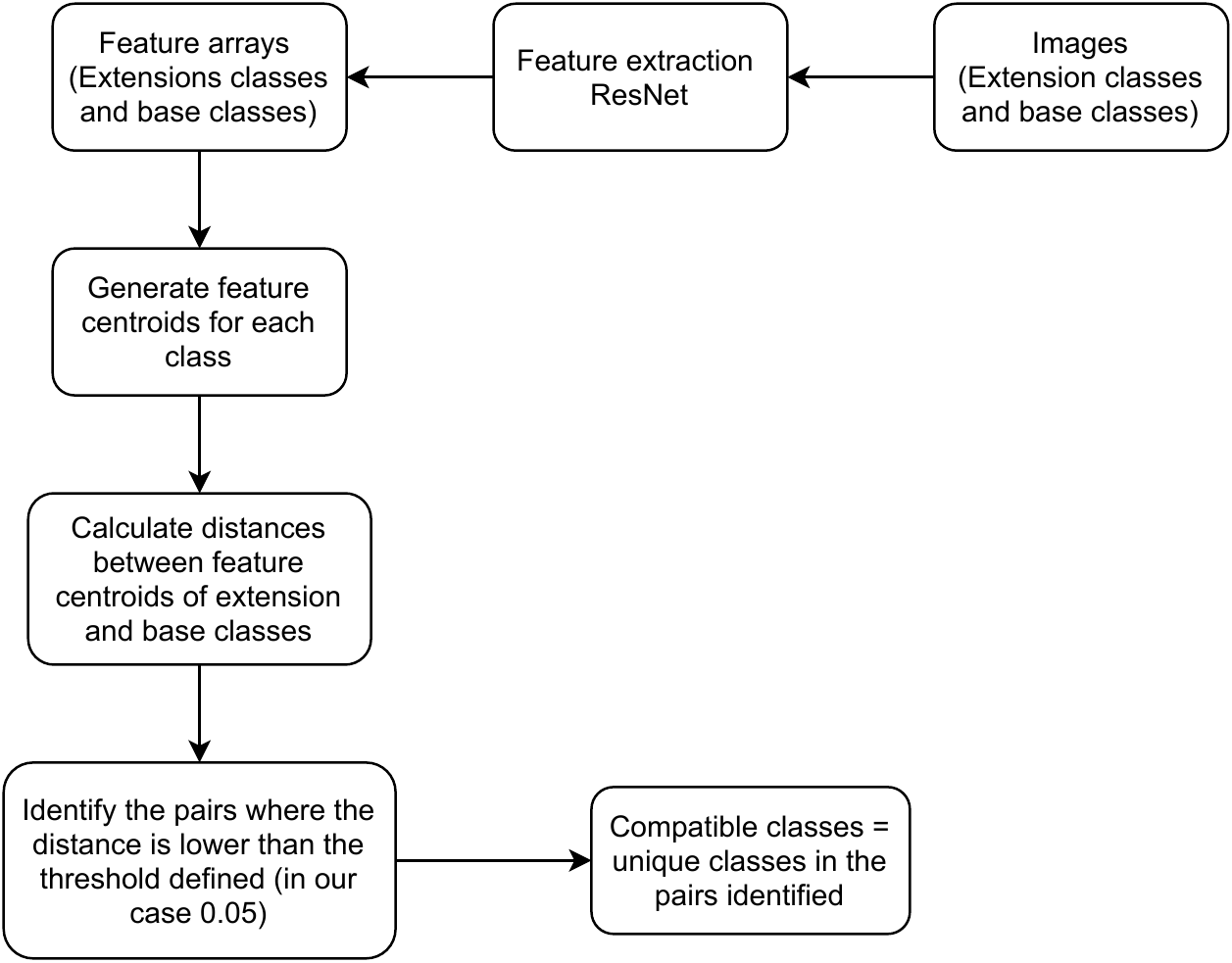}
\caption{Class compatibility calculation}
\label{figure1}
\end{figure}

The synopsis of the entire methodology can be viewed in Figure \ref{figure1}.

\subsubsection{Distance metric selection}

After trying various distance metrics with different layers, the combination of cosine and average pooling layer proved to be the most successful in identifying similar objects from the pool. Table-1 shows the resultant mMAP scores for the object similarity identification on the COCO dataset.

\begin{table}[ht]
\caption{Distance metric results.}
\begin{center}
\begin{tabularx}{\textwidth}{|X|X|X|X|}
\hline
\textbf{Model} & \textbf{Layer} & \textbf{Distance metric} & \textbf{MMAP} \\ \hline
ResNet152 & Average & Cosine & 0.7769 \\
ResNet152 & Average & D1 & 0.7847 \\
ResNet152 & Max pooling & D1 & 0.7480 \\
ResNet152 & Fully connected & Cosine & 0.7769 \\
ResNet152 & Fully connected & D1 & 0.7001 \\
ResNet152 & Fully connected & Square & 0.7011 \\
ResNet152 & Fully connected & D-Norm & 0.7101 \\ \hline
\end{tabularx}
\end{center}
\label{table:table1}
\end{table}

\subsubsection{Similarity threshold definition}

After various checks on various datasets like COCO, Oxford flowers, and Oxford IIIT dataset, the general value of the similarity threshold comes down to 0.05. Although the threshold can be adjusted according to the similarity of other classes in the dataset and the choice of the dataset itself.  
The scope for the S-E patch is defined in two ways. 
\begin{enumerate}
    \item Working without a tracking algorithm or tracking with live data. 
    \item Working with a tracking algorithm and recorded data.
\end{enumerate}

While working with the first case, the inference timings can be maintained by leveraging the use of tracking IDs and post-processing. In the second case, the system needs to go under parallelism to maintain the inference timings. Both the architectures for inference are mentioned in section \ref{inference1} and \ref{inference2}. 

\subsection{Dual-parallel inference technique} 
\label{inference1}

The dual-parallel inference technique is crucial to maintain the inference speed of the overall system. As mentioned earlier, this approach works on running both the models, detector and classifier, simultaneously but differing by one frame. The system revolves around the logic of sending the output of the detector to the classifier, which corrects and renders the output on the frame while the detector is processing the second frame. The shared memory consists of the frame and the prediction object. The prediction object is used to assess if there are any compatible classes there and if we need to run the classifier or directly render the frame.

Let us take an example to understand and further elaborate Algorithm \ref{algo:algo1p1} and Algorithm \ref{algo:algo1p2}.

\textbf{Algorithm \ref{algo:algo1p1}:} Imagine we have 80 base classes (refer to COCO class list) and we want to add a Van extension class. We first select the base classes highly similar to Vans by using the similarity threshold. In our case, these classes are Car, Bus, and Truck. It means that if there is a Van in the video, the model will trigger the region proposal and misclassify it as either of the three base classes. Hence we train a classifier on these four classes (three base classes and one extension class). We now can start the model inference.

\textbf{Algorithm \ref{algo:algo1p2}:} At time t0, the first frame is passed to the detection model and it gives a prediction object containing the detected classes, their bounding box coordinates, and their confidence scores. By looking at the classes, we can assess if the classifier needs to be invoked for that particular frame. If any of the three classes are predicted, we need to run the classifier in that region of the frame. The prediction object and frame are copied into the shared memory, and the lock is released.

\begin{figure}[!t]
\centering
\includegraphics[width = 0.9\columnwidth,scale=0.2]{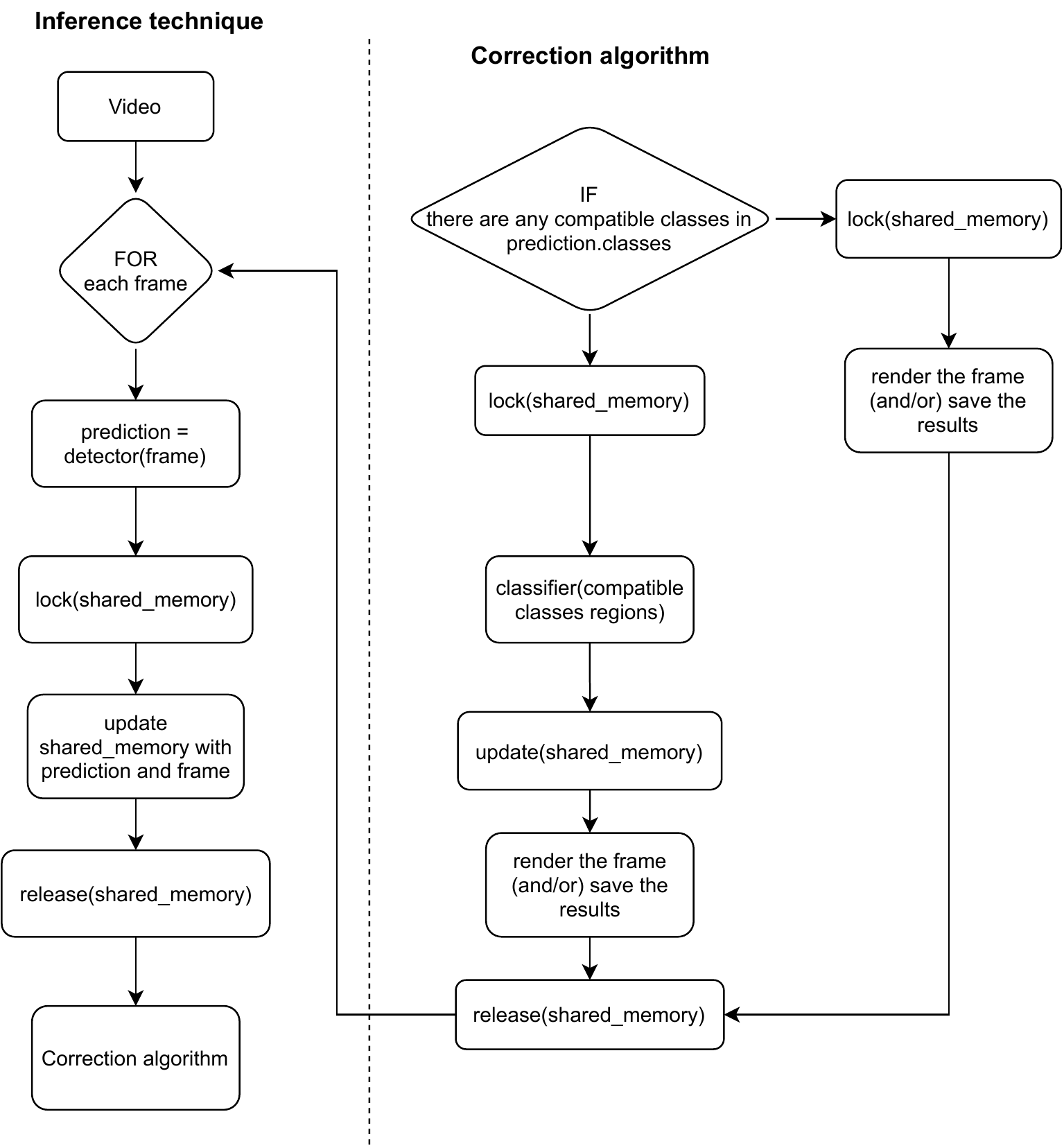}
\caption{Dual-parallel inference methodology}
\label{figure2}
\end{figure}

The classifier locks and accesses the shared memory and corrects the prediction object with a new class (depending on the classifier's output) and renders the frame after releasing the lock at time t1. The shared memory is then updated once the detector, which has already processed the second frame, gets a lock on it, and the process is repeated till time tn. The synopsis of the entire methodology can be viewed in Figure \ref{figure2}.

\subsection{Technique with trackers}
\label{inference2}

I also have tried the technique with tracker algorithms which is one of the best applications of the procedure. Whenever we use tracking algorithms with object detectors, we get better approaches to improve results, basically due to the option of remembering each object with a unique ID. This allows us to save each object ID at its best size and then run correction algorithms. I used the same technique and extracted the ID whose classes matched with the compatible classes. The best thing is it doesn't take that much space or time in reprocessing either. The time is dropped to just the number of objects with the class matching to any of the compatible classes.

\subsection{Model architectures}

I have used the traditional ResNet152 classification architecture to train similar classes. It was pre-trained on ImageNet and used the standard configuration with Cross entropy loss and step learning rate scheduler. The object detection model I used was the Yolov5x by Ultralytics. The technique with trackers utilizes the DeepSort algorithm by N Wojke.

\subsection{Training environment }
All the models were trained on Intel Xeon E5-1620 v3 (@ 3.50GHz × 8 CPU)  with Nvidia GeForce RTX2080Ti GPU and 16GB of RAM. I used the standard PyTorch library to code and train the model architectures. All models were trained with CUDA enabled. \footnote{All code can be accessed at-https://github.com/Dishant-P/S-Extension-patch-Official-research-module.git}

%<--------Algorithm1------->

\begin{algorithm}
\textbf{INPUT:}\\$features (n_{c} \times n_{i} \times 2048)$: List of feature arrays of every class image in the database (base + extension classes), $threshold (float)$: distance threshold to consider a class as compatible\\
\textbf{OUTPUT:}\\$compatible{\_}classes (n)$: List of compatible classes\\

\begin{algorithmic}
\STATE \textbf{Function} similarity{\_}compatibility($dataset$, $threshold$, $extension{\_}class{\_}list$, $base{\_}class{\_}list$):
\STATE \hspace{0.3175cm} $classes \leftarrow extension{\_}class{\_}list$ + $base{\_}class{\_}list$
\STATE \hspace{0.3175cm} $i \leftarrow 0$
\STATE \hspace{0.3175cm}\textbf{for} length of $f{\_}list$ do

\STATE \hspace{0.635cm}$class{\_}features \leftarrow features[classes[i]]$ 
\STATE \hspace{0.635cm}$model \leftarrow KMeans(n{\_}clusters = 1)$
\STATE \hspace{0.635cm}$model.fit(class{\_}features)$
\STATE \hspace{0.635cm}$feature{\_}centroids[classes[i]] \leftarrow model.cluster{\_}centers{\_}$
\STATE \hspace{0.635cm}$i \leftarrow i + 1$

\STATE \hspace{0.5cm}
\STATE \hspace{0.3175cm} $i \leftarrow 0$
\STATE \hspace{0.3175cm}\textbf{for} length of $extension{\_}class{\_}list$ do
\STATE \hspace{0.635cm} $j \leftarrow 0$
\STATE \hspace{0.635cm}\textbf{for} length of $base{\_}class{\_}list$ do
\STATE \hspace{0.9525cm}$distance \leftarrow cosine(feature{\_}centroids[extension{\_}class{\_}list[i]],$
\STATE \hspace{5.27cm}$feature{\_}centroids[base{\_}class{\_}list[j]])$
\STATE \hspace{0.9525cm} \textbf{if} $distance$ \textgreater 0.05:
\STATE \hspace{0.9525cm} $compatible{\_}classes[extension{\_}class{\_}list[i]] \leftarrow [base{\_}class{\_}list[j]]$
\STATE \hspace{0.9525cm}$j \leftarrow j + 1$
\STATE \hspace{0.635cm}$i \leftarrow i + 1$
\STATE \hspace{0.3175cm}\textbf{return} $compatible{\_}classes$

\end{algorithmic}
\caption{Algorithm for finding compatible classes for training classifier}
\label{algo:algo1p1}
\end{algorithm}

%<----------Algorithm2-------->

\begin{algorithm}
\textbf{INPUT:} $video$: The inference video\\
\textbf{OUTPUT:} Prediction
\begin{algorithmic}
\STATE \hspace{0.3175cm}\textbf{for} length of $video$
\STATE \hspace{0.635cm}$prediction \leftarrow detector(frame)$
\STATE \hspace{0.635cm}$lock(shared{\_}memory)$
\STATE \hspace{0.635cm}update $shared{\_}memory$ with prediction and frame
\STATE \hspace{0.635cm}$release()$
\STATE \hspace{0.635cm}\textbf{if} there are any $compatible{\_}classes$ in $prediction.classes$
\STATE \hspace{0.9525cm}$lock(shared{\_}memory)$ 
\STATE \hspace{0.9525cm}$classifier(compatible{\_}classes{\_}$ regions $)$
\STATE \hspace{0.9525cm}update $(shared{\_}memory)$
\STATE \hspace{0.9525cm}render the frame (and/or) save the results
\STATE \hspace{0.635cm}\textbf{else}
\STATE \hspace{0.9525cm}$lock(shared{\_}memory)$
\STATE \hspace{0.9525cm}render the frame (and/or) save the results
\STATE \hspace{0.9525cm}$release()$
\end{algorithmic}
\caption{Dual-parallel inference algorithm}
\label{algo:algo1p2}
\end{algorithm}

\section{Experiments and Discussions}
\label{Experiments and Discussions}
The S-Extension patch outperforms the previous techniques in terms of the following criteria:
\begin{enumerate}
    \item Speed and ease of adaptation 
    \item Less need of data at the instance 
    \item Adaptability to any network or system 
    \item Computational resources required 
\end{enumerate}

Additionally, the S-Extension patch can maintain the accuracy of the overall model as well. The S-Extension patch breaks down the speed of training and computational resources required by at least 75{\%}. 

In the fine-tuning technique, the model forgets the previous knowledge and hence is not able to perform the previous tasks properly. The joint training technique works well on both old and new tasks but is not able to train quickly and requires a lot of data and annotations for the same. The technique of Learning without forgetting closes off both the previous limitations for the need of data and performance, but it still takes the same time for training as a regular object detection model and has the need for annotating the new data as well. Also, all three techniques require significant computational power to train the model. 

\begin{table}
\caption{Comparison of S-Extension patch with earlier techniques}
\begin{center}
\begin{tabularx}{\textwidth}{|X|X|X|X|X|X|}
\hline
\textbf{Technique} & \textbf{Need for data preprocessing} & \textbf{Speed of adaptation} & \textbf{Time for training} & \textbf{Performance on old tasks} & \textbf{Performance on extension tasks} \\ \hline
Learning without forgetting & Low & Slow & Medium & Best & Best \\
Fine tuning & Medium & Fast & Medium & Good & Fair \\
Joint training & High & Fast & High & Best & Best \\
S-Extension patch & None & Fast & Low & Best & Best \\ \hline
\end{tabularx}
\label{table:table4}
\end{center}
\end{table}

\begin{table}[ht]
\caption{Similarity matrix for three classes from COCO dataset and an extension class.}
\begin{center}
\begin{tabularx}{\textwidth}{|X|X|X|X|X|}
\hline
\textbf{Class} & \textbf{Bus} & \textbf{Car} & \textbf{Truck} & \textbf{Van} \\ \hline
Bus & 0 & 0.0977 & 0.0314 & 0.0468 \\
Car & 0.977 & 0 & 0.0685 & 0.0378 \\
Truck & 0.0314 & 0.0685 & 0 & 0.0292 \\
Van & 0.0468 & 0.0378 & 0.0292 & 0 \\ \hline
\end{tabularx}
\label{table3}
\end{center}
\end{table}

The S-Extension patch addresses all these limitations by leveraging the knowledge of the older tasks. I do concede there is some need for old tasks data but from an industry point of view, the old data is not a problem but the preprocessing time and cost of training is. As the S-Extension patch only involves training a regular classifier, it doesn’t involve many preprocessing steps, especially annotation. And it also doesn’t need the computational resources or time for training. Table \ref{table:table4} sums up the entire argument on the criteria. Note: all techniques have relatively similar inference timings. 

\subsection{Dual-parallel inference technique} \label{dual}

In the experiments, I trained a YOLOv5x model over the COCO dataset using a pre-trained set. The model gave a test mMAP (0.5:0.95) score of 50.4. The extension class: Van, was appended to the 80 base classes of the COCO dataset. In the joint training method, it took about 120 hours for my system to train the model and achieve the overall test mMAP (0.5:0.95) score of 49.8 (with extension class appended). In the S-Extension patch method, however, it took 2.4 hours for a ResNet152 classifier model to reach mMAP (0.5:0.95) score of 50.7 for the four similar classes namely, Car, Bus, Truck and Van (extension class), identified with the similarity threshold, as shown in Table \ref{table3}.  

The increase in the mMAP score for the S-Extension patch is not only because of the higher accuracy of extension classes but also an increase in the classification accuracy of the other three base classes. As seen in Table \ref{table3}, not only Van but even Truck class is highly similar with Bus. It means that by training the classifier for highly similar classes, the model gets an additional chance to correct the classification done by the detection model and increase the overall accuracy. It is one of the most crucial advantages of the S-Extension patch as it allows us to use this technique not only in need of extension classes but also when a particular class gets a lower score or while working with a dataset with highly similar classes. 

\subsection{Technique with trackers}

In the tracker-based algorithm technique, I used the DeepSort algorithm with class labels extraction. Although usually, the DeepSort tracking algorithm is not used with class labels, it was important to extract those to use it with the S-Extension patch. The tracker algorithm-based approach has one crucial advantage about the overall accuracy. Not only does it help in increasing the overall accuracy like the one explained in Section \ref{dual}, but also in reducing the additional computational resources required and the inference timings. The technique with trackers achieves the same final mMAP score (0.5:0.95) of 50.7 as with the Dual-parallel inference technique but with 12{\%} lower inference timings and 37{\%} decrease in the overall computational usage.  

\section{Future works and limitations}
Although the S-Extension patch covers most of the limitations, it is a relatively new technique with a lot of scope still to be explored. The technique can be further improved by using other similarity measures or appending the patch directly to the model while training and keeping a unified architecture. The S-Extension patch can be made more scalable by using a two-step detection model and using the patch for just the classification step, and fine-tuning the model there. It can significantly improve the speed and the time to implement the inference strategy. The strategy is also being experimented on on different visual models including classifiers and graphics rendering algorithms.

\section{Conclusion}
The paper shows how we can leverage the knowledge of the previous data and performance to quickly extend the object detection model and implementation variations of the technique. I give considerable evidence to prove how the technique works better than other defined strategies on various criteria. The S-Extension patch is an elegant technique that can be used with any object detection model fulfilling the similarity threshold condition. S-Extension patch is easy to adapt to any system, can be trained quickly, and maintains (and in some cases increases) the accuracy as well the inference speed. 

\bibliographystyle{splncs04}
\bibliography{spl-template}

\end{document}